\documentclass[sigconf,nonacm,pbalance=true]{template/acmart}
\usepackage{array}
\usepackage{colortbl}
\usepackage{multirow}
\usepackage{threeparttable}
\usepackage[ignoredisplayed]{enumitem}
\setcopyright{none}
\renewcommand{\footnotetextcopyrightpermission}[1]{}
\newcommand{\bench}{DuSE-Bench}

\newcommand{\policy}[1]{\textbf{#1}}
\newcommand{\pct}{\%}

\makeatletter
\def\@ACM@checkaffil{}
\makeatother
\begin{document}
\title{DuSE-Bench: Benchmarking Joint Reasoning over Evolving Dialogue and Documents}
\author{Guanjie Cheng\texorpdfstring{\textsuperscript{1,2}}{} \quad Jianfeng Gao\texorpdfstring{\textsuperscript{2}}{} \quad Yufei Tong\texorpdfstring{\textsuperscript{3*}}{} \quad Xinkui Zhao\texorpdfstring{\textsuperscript{1,2}}{} \quad Shuiguang Deng\texorpdfstring{\textsuperscript{3*}}{}}
\affiliation{\vspace{4pt}\institution{\textsuperscript{1}Ningbo Global Innovation Center, Zhejiang University}}
\affiliation{\institution{\textsuperscript{2}School of Software Technology, Zhejiang University}}
\affiliation{\institution{\textsuperscript{3}College of Computer Science and Technology, Zhejiang University}}
\renewcommand{\shortauthors}{Cheng et al.}
\begin{abstract}
Persistent Web information assistants must combine evolving document facts with user requirements established and revised across sessions. Yet systematic evaluation of how changes in both sources affect response requirements and evidence attribution remains limited. To address this gap, we introduce Dual-Source Evolution Benchmark \textbf{(DuSE-Bench)}, a benchmark comprising 1,000 tasks across five scenarios, including 700 whose required premises originate in both dialogue and documents. An evolving Support Graph guides task construction and preserves links from updated facts and reused conclusions to original turns and passages. Task annotations distinguish grounding source from state dependence and enable separate evaluation of response content, response type, and source attribution. Experiments across 11 baseline configurations reveal persistent challenges in joint reasoning over evolving dialogue and documents, with content accuracy on jointly grounded tasks reaching at most 56.7\%. Moreover, improvements in content accuracy can coincide with poorer response choice, while content-correct responses frequently lack complete attribution to original evidence. Matched evidence substitutions further show that the effect of replacing one source's evidence depends on support from the other. These findings motivate coordinating dialogue memory and document retrieval to apply current user requirements and document facts together while preserving their original support.
The data are available at \url{https://github.com/JianfengGao7/DuSE-Bench}.
\end{abstract}
\maketitle
\begingroup
\renewcommand{\thefootnote}{\fnsymbol{footnote}}
\footnotetext[1]{Corresponding authors.}
\endgroup
\hypersetup{pdfauthor={Guanjie Cheng, Jianfeng Gao, Yufei Tong, Xinkui Zhao, Shuiguang Deng}}
\pdfinfo{/Author (Guanjie Cheng, Jianfeng Gao, Yufei Tong, Xinkui Zhao, Shuiguang Deng)}
\section{Introduction}
Persistent Web information assistants support continuing work such as news analysis and software maintenance by combining information gathering~\cite{browsecomp} with memory across conversations~\cite{longmemevalv2,memoryos,secom}. As work progresses, users may revise their requirements, while documents accumulate and their facts change across versions or successive reports~\cite{newsedits,re3}. Assistants must therefore determine which user requirements and document facts apply to the current request and how they should be used together.

Existing benchmarks evaluate aspects of these capabilities, including long-context understanding~\cite{longbench}, memory updating and interaction~\cite{longmemeval,memoryagentbench,amemgym,memoryarena}, version-specific document use~\cite{versionrag,tide}, and joint use of dialogue and documents~\cite{memorydocdataset,rhelm}, as Table~\ref{tab:benchmark_comparison} shows. However, systematic evaluation of how assistants combine applicable information from evolving dialogue and documents, choose appropriate responses, and cite original supporting evidence remains limited. This setting presents three connected challenges.

\begin{table*}[t]
\centering
\captionsetup{skip=3pt}
\caption{Comparison of DuSE-Bench with related benchmarks. Turns are mean message counts per original session. Token counts (K) are per document or released source unit, reported as approximate means ($\sim$) or ranges. \ding{51} and \ding{55} indicate whether a benchmark meets the capability criteria; -- denotes not applicable. Definitions and sources are provided in Appendix~\ref{sec:comparison_definitions}.}
\label{tab:benchmark_comparison}
\begingroup
\small
\setlength{\tabcolsep}{3pt}
\renewcommand{\arraystretch}{1.02}
\setlength{\aboverulesep}{0.3ex}
\setlength{\belowrulesep}{0.45ex}
\newcommand{\capyes}{\textcolor[HTML]{00B000}{\ding{51}}}
\newcommand{\capno}{\textcolor[HTML]{F00000}{\ding{55}}}
\begin{tabular}{w{l}{80pt}w{c}{23pt}w{c}{41pt}w{c}{41pt}w{c}{50pt}w{c}{48pt}w{c}{42pt}w{c}{56pt}w{c}{\dimexpr0.95\textwidth-381pt-18\tabcolsep\relax}}
\toprule
\multirow{2}{*}[-6pt]{\textbf{Benchmark}} & \multicolumn{3}{c}{\textbf{Data characteristics}} & \multicolumn{5}{c}{\textbf{Capabilities}} \\
\cmidrule(lr){2-4}\cmidrule(lr){5-9}
 & \raisebox{4pt}{\textbf{Turns}} & \raisebox{4pt}{\textbf{Tokens (K)}} & \raisebox{4pt}{\textbf{Domains}} & \textbf{\shortstack{Cross-session\\dialogue}} & \textbf{\shortstack{Evolving\\documents}} & \textbf{\shortstack{Joint\\reasoning}} & \textbf{\shortstack{Response-type\\evaluation}} & \textbf{\shortstack{Source\\attribution}} \\
\midrule
LoCoMo & 21.6 & -- & Daily life & \capyes & \capno & \capno & \capyes & \capno \\
LongMemEval & 10.4 & -- & Personal & \capyes & \capno & \capno & \capyes & \capno \\
VersionQA & -- & $\sim$13.4 & Software & \capno & \capyes & \capno & \capno & \capno \\
TIDE & -- & $\sim$3.1 & Law & \capno & \capyes & \capno & \capno & \capno \\
mtRAG & 15.3 & $\sim$0.3 & Multiple & \capno & \capno & \capyes & \capyes & \capno \\
MemoryDocDataSet & 10.6 & 20--50 & Law & \capyes & \capno & \capyes & \capyes & \capno \\
RHELM & 37.4 & $\sim$1.5 & Multiple & \capyes & \capyes & \capyes & \capno & \capno \\
\midrule
\rowcolor[HTML]{F0F4F8}
\textbf{DuSE-Bench (ours)} & \textbf{37.1} & \textbf{2--420} & \textbf{Multiple} & \capyes & \capyes & \capyes & \capyes & \capyes \\
\bottomrule
\end{tabular}
\endgroup
\vspace{-5pt}
\end{table*}

\begin{figure}[t]
  \centering
  \includegraphics[width=0.9\columnwidth]{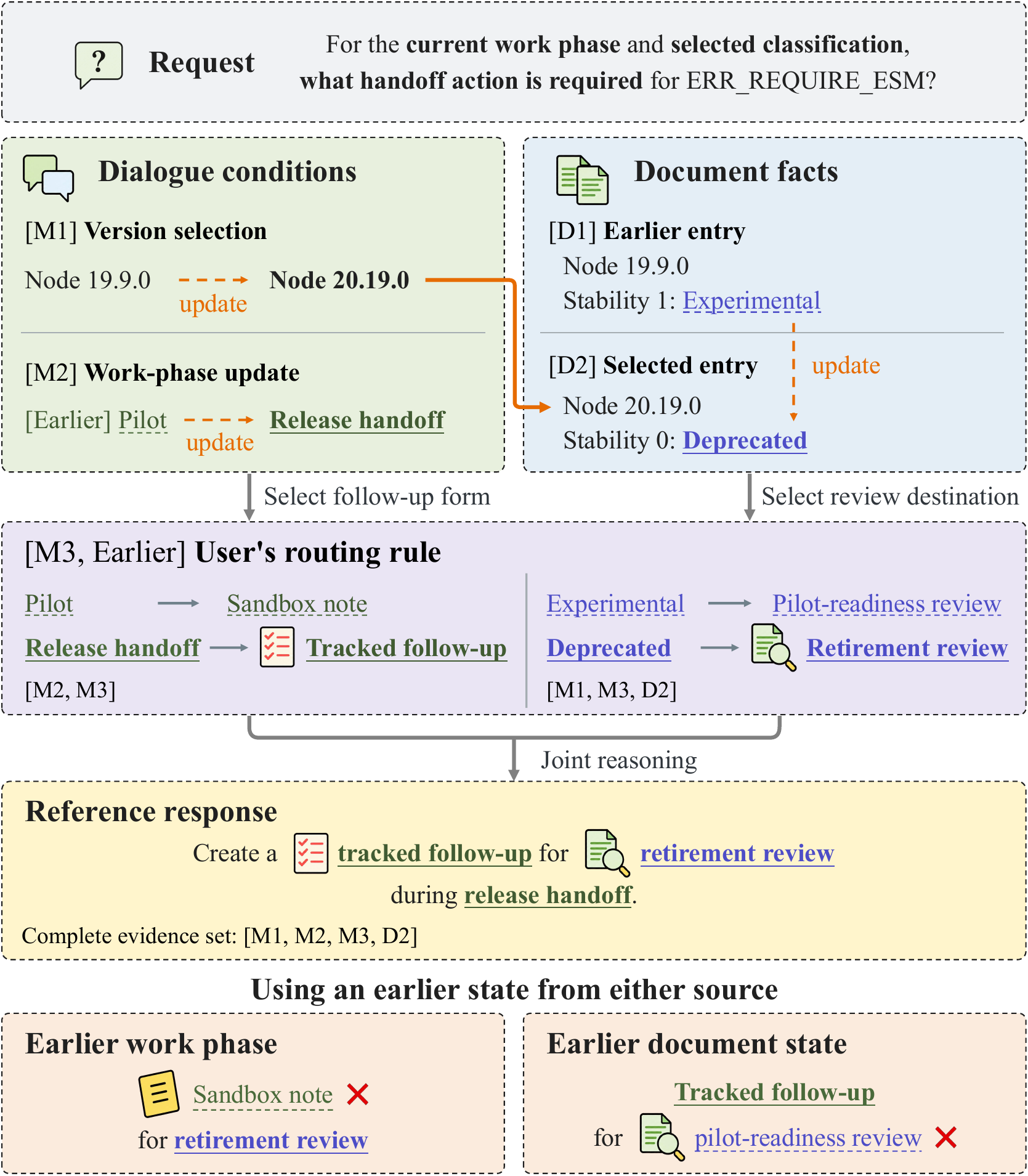}
  \vspace{-8pt}
  \Description{A software-release request combines dialogue conditions with versioned document facts. The selected Node 20.19.0 entry classifies the error as Deprecated, and the current work phase is release handoff. The routing rule requires a tracked follow-up for retirement review. Restoring the earlier work phase changes the follow-up form with the document classification fixed. Restoring the earlier document classification changes the review destination with dialogue conditions fixed.}
  \caption{Joint reasoning over evolving dialogue and documents in a software-release task. Restoring either source's earlier state changes the handoff action while the other remains fixed.}
\vspace{-8pt}
  \label{fig:dialogue-example}
\end{figure}

\textbf{Joint reasoning over evolving sources.}
A request may require document facts to be interpreted under user conditions established earlier in the dialogue. In the software-release task in Figure~\ref{fig:dialogue-example}, an earlier user rule determines the handoff action from the selected document's error classification and the work phase recorded in dialogue. Under this rule, changing the document classification alters the review destination, while changing the work phase alters the follow-up form. More generally, a request may require both sources even when only one has relevant changes, because necessary facts or conditions from the other still apply.

\textbf{Response choice under evolving conditions.}
Even when relevant facts are available, a request may depend on a user choice left unresolved in the dialogue~\cite{sharc,rhelm}. In news analysis, for example, documents may contain several speakers' statements, but the dialogue may not establish whose view the user wants summarized. Selecting a speaker would then treat an unresolved user choice as settled, so the assistant should ask whose view is intended. Other unresolved issues call for different responses: conflicts between applicable claims require explanation, while factual gaps require stating any supported result and identifying missing evidence. Assistants must therefore track unresolved issues as work progresses and choose responses accordingly.

\textbf{Traceable support for responses.}
Assistants often reuse document facts and earlier conclusions when revising reports or supporting decisions. This carries useful information forward, but the links to original passages and dialogue turns may not be retained. Users may then struggle to verify a reported fact or identify the conditions supporting a reused conclusion. Even when such reuse produces correct content, the response may still lack complete attribution to original evidence. This motivates evaluating whether citations identify the original turns and passages that jointly justify the required response under the applicable conditions.

To address these challenges, we introduce Dual-Source Evolution Benchmark \textbf{(DuSE-Bench)}, a benchmark for joint reasoning over evolving dialogue and documents in multi-session workflows. We construct dialogue and tasks around published document versions or reports, incorporating changes in user requirements and reuse of earlier results. An evolving Support Graph records user conditions, document facts, and their updates, linking tasks and earlier conclusions to their supporting information. The accumulated graph guides subsequent session and task design while preserving links from tasks to original turns and passages.

The resulting benchmark contains 1,000 tasks across five scenarios, including 700 whose required premises originate in both dialogue and documents. Task annotations distinguish grounding source from state dependence to support performance comparisons across both dimensions. Each task includes reference content, a required response type, and a complete evidence set for separate evaluation of response content, response type, and source attribution. Experiments with 11 baseline configurations and controlled evidence interventions reveal persistent difficulty in using dialogue and document support together.

In summary, our main contributions are:
\begin{itemize}[leftmargin=1.5em, topsep=2pt, partopsep=1pt]
  \item \textbf{Benchmark and task design.}
  We introduce \bench{} for joint reasoning over evolving dialogue and documents in multi-session workflows, with task categories separating source requirements from state dependence.

  \item \textbf{Traceable construction and evaluation.}
  We develop a Support Graph-guided construction protocol that preserves links from tasks to original evidence through updates and reuse. Reviewed annotations enable separate evaluation of response content, response type, and source attribution.

  \item \textbf{Empirical findings.}
  Experiments reveal persistent challenges in joint reasoning over evolving dialogue and documents. Improvements in content accuracy can coincide with poorer response choice, while content-correct responses frequently lack complete attribution to original evidence.
\end{itemize}

\section{Related Work}
\label{sec:related_work}

\subsection{Memory updates and document applicability.}
Long-term memory benchmarks such as LoCoMo evaluate recall and reasoning over information accumulated across sessions~\cite{locomo}. When later statements revise or invalidate earlier information, a response must use the facts that apply to the current request. LongMemEval tests knowledge updates by requiring responses to use revised information~\cite{longmemeval}. Memora examines repeated revisions and invalidations, explicitly rewarding correct use of valid facts and penalizing reliance on obsolete information~\cite{memora}.

For document evidence, applicability likewise depends on the requested version or period, so an earlier text may be the correct source, a concern also reflected in time-sensitive QA benchmarks~\cite{timeqa,freshllms,crag}. VersionQA tests version-specific content and changes in software documentation~\cite{versionrag}. For date-specific requests, TIDE requires identifying the rule in force on the query date, including when the supplied version does not apply~\cite{tide}. When earlier dialogue specifies the relevant version or date, applying document facts requires relating them to that condition.

\subsection{Joint use of dialogue and documents.}
Dialogue can therefore guide passage retrieval and the use of document facts~\cite{doc2dial,multidoc2dial}. In mtRAG, earlier turns clarify history-dependent questions and guide retrieval of the passages needed for an answer~\cite{mtrag}. User information also guides passage retrieval and answer generation in iKAT through a supplied personal text knowledge base~\cite{ikat2024}. Missing user conditions can also affect response choice: ShARC applies rule text to user circumstances and follow-up history to decide whether to answer or seek clarification~\cite{sharc}.

Long-term joint benchmarks connect retained dialogue with external documents across sessions (Table~\ref{tab:benchmark_comparison}). MemoryDocDataSet's Hybrid questions use retained history to identify a relevant document and obtain the answer from its contents~\cite{memorydocdataset}. RHELM synchronizes external sources with evolving user trajectories, and its Mixed tasks include analysis after document modifications~\cite{rhelm}. Its dialogue-based Misleading tasks also require recognizing requests that conflict with updated user states and proposing suitable alternatives. \bench{} distinguishes source requirements from state dependence, checking which sources are necessary and whether restoring a recorded prior state alters the required response.

\subsection{Evidence and response evaluation.}
Evaluating a response also involves checking whether the evidence provided is sufficient to justify it~\cite{hotpotqa,tong2026longdocvqa}. For supported or refuted claims, FEVER's joint score requires both the correct label and a complete annotated evidence set~\cite{fever}. For generated answers, ALCE evaluates citation support for individual statements separately from answer correctness~\cite{alce}. Its analyses show that well-supported statements can still leave the question incompletely answered. Across sessions, restated document facts and reused conclusions can obscure which original sources justify a response. \bench{} pairs reviewed response requirements with complete evidence sets of original turns and passages, supporting separate evaluation of content, response type, and source attribution.

\begin{figure*}[t]
  \centering
  \includegraphics[width=0.96\textwidth]{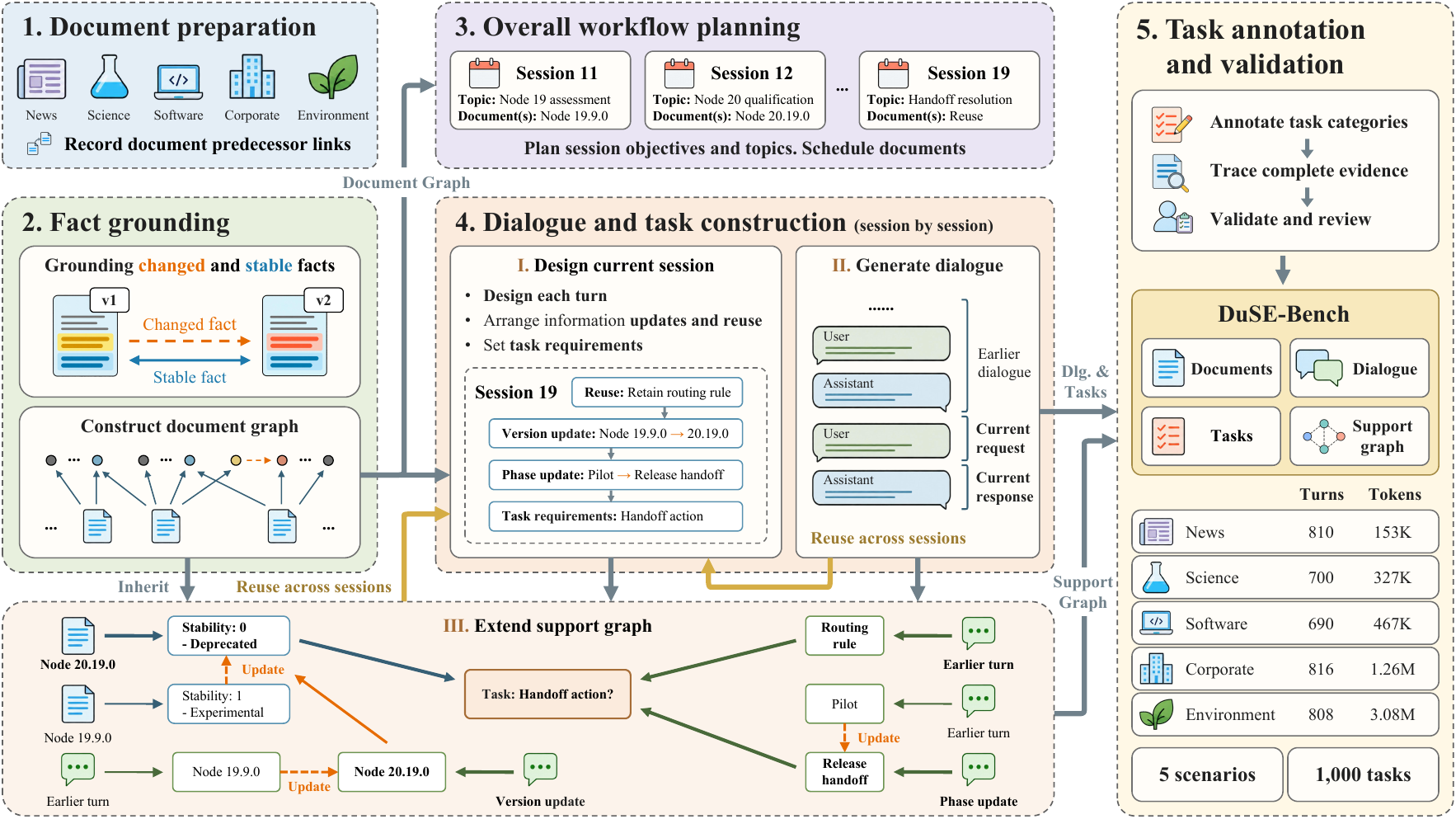}
  \Description{Five panels connect original document collections, the Document Graph, a workflow plan, dialogue and task construction, and task annotation and validation. Document preparation preserves source text and links related versions, reports, or assessment stages in their documented order. The Document Graph grounds facts in original passages and records changes across linked documents. The plan assigns work goals, document introductions, and task coverage across sessions. Reviewed session designs guide dialogue generation, using earlier dialogue and the accumulated Support Graph. The Support Graph traces task dependencies to original turns and passages, retaining support as conditions change. The final panel summarizes task annotation and human review and shows the benchmark's documents, dialogue, annotated tasks, and Support Graph, with counts of dialogue turns and tokens across documents and dialogue.}
  \caption{Construction workflow of \bench{}. The evolving Support Graph guides session and task design. It preserves links to original evidence through updates and reuse to support task annotation and validation.}
  \label{fig:dataset-construction}
\vspace{-5pt}
\end{figure*}

\section{Task Definition and Evaluation}
\label{sec:task}

In this section, we first define the evolving dialogue and document environment in Section~\ref{sec:task_context} and characterize task dependencies in Section~\ref{sec:task_categories}. Building on these definitions, Section~\ref{sec:responses_and_evidence} specifies response requirements and task references. Finally, Section~\ref{sec:evaluation} presents the protocol for separately evaluating response content, response type, and source attribution.

\subsection{Evolving dialogue and documents}
\label{sec:task_context}
Each scenario represents a multi-session workflow in one domain. Recorded dialogue $H$ contains user conditions and earlier conclusions that later requests may reuse. Documents from a collection $D$ are introduced as the work progresses and remain available in subsequent sessions. Users may revise conditions or choices, while document facts may change across versions or successive reports. These changes can occur independently.

Within each session, a \textbf{turn} is one user or assistant message, and a \textbf{round} pairs a user turn with the following assistant turn. At user turn $t$, request $q_t$ defines a task $\tau$ in the context of preceding dialogue $H_{<t}$ across sessions and documents $D_{\leq t}\subseteq D$ introduced by that turn. The request's version, time, and scope requirements, together with recorded updates, determine which information applies. For example, a request about an earlier version may require its document even after a later version is introduced.

\subsection{Task dependencies}
\label{sec:task_categories}
\textbf{Grounding source} identifies whether the premises needed to resolve a task's underlying request originate in earlier dialogue, documents, or both. Dialogue-grounded and jointly grounded tasks require prior user information that the current request does not fully supply. Classification follows original sources: a document fact retains its origin when restated in dialogue or stored in memory.

\textbf{State dependence} identifies which recorded changes to these premises affect the required response. A change is relevant if restoring its prior state would alter the required result, a necessary qualification, or the response type. Static tasks have no necessary premise with a relevant recorded change. Dialogue and document dependence indicate relevant changes in the corresponding source alone. Dual dependence applies when restoring either source's earlier state changes the required response with the other source held fixed. Thus, a jointly grounded task can be static or depend on changes in one or both sources.

Among jointly grounded tasks, \textbf{integration structure} describes how premises from the two sources are combined. Direct tasks combine dialogue and document premises without requiring a prior conclusion derived within one source. In chained tasks, an earlier conclusion follows from at least two jointly necessary premises of one source type and is then combined with premises from the other. The later task may reuse this conclusion or reconstruct it from its original premises. Appendix~\ref{sec:worked_evidence_paths} illustrates a chained task and traces its original support.

\begin{table}[t]
\centering
\captionsetup{font=normalsize}
\caption{Task dimensions and response types in \bench{}.}
\label{tab:task_dimensions}
\begin{threeparttable}
\begingroup
\small
\setlength{\tabcolsep}{4pt}
\renewcommand{\arraystretch}{1.05}
\setlength{\aboverulesep}{0.3ex}
\setlength{\belowrulesep}{0.45ex}
\begin{tabular}{ll}
\toprule
\rowcolor[HTML]{F0F4F8}
\textbf{Dimension} & \textbf{Categories} \\
\midrule
Grounding source & Dialogue, Document, Joint \\
State dependence & Static, Dialogue, Document, Dual \\
Integration structure & Direct, Chained \\
\midrule
Response type & \begin{tabular}[c]{@{}l@{}}Answer, Clarification request,\\Conflict report, Abstention\end{tabular} \\
\bottomrule
\end{tabular}
\endgroup
\end{threeparttable}
\end{table}

\subsection{Responses and task references}
\label{sec:responses_and_evidence}
The request and its applicable premises determine the required response content and type. When the request is clear and sufficiently supported, an \policy{answer} gives the result and its necessary qualifications. A \policy{clarification request} asks the user to supply an essential missing condition or resolve an ambiguous reference left unsettled by the dialogue. A \policy{conflict report} explains incompatible applicable claims about the same fact that remain unresolved after accounting for version, time, scope, and stated priorities. For a clear request with missing factual support, an \policy{abstention} response gives any supported result, identifies the missing evidence, and withholds unsupported conclusions.

A \textbf{complete evidence set} consists of original turns or document passages available at $q_t$ that jointly justify the required response, with every entry necessary. It includes recorded updates when they are needed to establish which premises apply. Each task is annotated with reference content, the required response type, and a default complete evidence set $C_t^0$. Each accepted alternative must independently meet the same criterion for the same response and request conditions. Section~\ref{sec:validation_assembly} describes their annotation and validation.

We represent a model response as
\begin{equation}
  y_t=(a_t,\pi_t,C_t),
\end{equation}
where $a_t$ is the response text, $\pi_t$ is the predicted response type, and $C_t$ is the set of cited original turn and passage identifiers.

\subsection{Evaluation protocol}
\label{sec:evaluation}
We score response content $a_t$, response type $\pi_t$, and cited support $C_t$ separately to distinguish task completion, response choice, and source attribution.

\textbf{Content.}
We use LLM-as-Judge~\cite{mtrag,longmemeval} to assess response content $a_t$ through two scores: content accuracy and content quality. Content accuracy is the fraction of tasks whose responses correctly supply every essential result and qualification without material error. Content quality credits partial completion through an independently assigned 0--5 grade, divided by five before averaging. Appendix~\ref{sec:metric_details} details judging and scoring rules.
\textbf{Response type.}
We compare $\pi_t$ with the required response type and report Macro-F1, giving the four types equal weight despite their unequal frequencies.
\textbf{Source attribution.}
Source attribution F1 compares cited identifiers $C_t$ with those of a single complete reference set $C_t^{\star}$, chosen from $C_t^0$ and accepted alternatives to maximize F1.

\section{Benchmark Construction}
\label{sec:benchmark}
Figure~\ref{fig:dataset-construction} outlines our construction process. We first prepare documents and ground facts (Sections~\ref{sec:document_preparation}--\ref{sec:document_graph}), then plan the workflow (Section~\ref{sec:workflow_planning}). Dialogue and task construction extends the Support Graph across sessions (Section~\ref{sec:dialogue_task_construction}), followed by task annotation and validation (Section~\ref{sec:validation_assembly}).

\begin{table*}[t]
\centering
\caption{DuSE-Bench statistics by scenario. Turns count individual messages. Token totals (K) cover document and dialogue text.}
\label{tab:dataset}
\small
\begin{tabular}{llrrrrrr}
\toprule
\textbf{Scenario} & \textbf{Domain} & \textbf{Documents} & \textbf{Passages} & \textbf{Sessions} & \textbf{Turns} & \textbf{Tasks} & \textbf{Tokens (K)} \\
\midrule
1 & News & 38 & 211 & 17 & 810 & 200 & 153 \\
2 & Science & 30 & 442 & 18 & 700 & 200 & 327 \\
3 & Software & 22 & 781 & 20 & 690 & 200 & 467 \\
4 & Corporate & 22 & 1,998 & 20 & 816 & 200 & 1,255 \\
5 & Environment & 14 & 3,943 & 28 & 808 & 200 & 3,081 \\
\midrule
\rowcolor[HTML]{F0F4F8}
Total & Five domains & 126 & 7,375 & 103 & 3,824 & 1,000 & 5,283 \\
\bottomrule
\end{tabular}
\end{table*}

\subsection{Document preparation}
\label{sec:document_preparation}
The document collection $D$ contains published versions and reports across News, Science, Software, Corporate, and Environment. These documents come from NewsEdits-BBC, Re3-Sci, VersionQA, LEDGER, and NEPATEC, respectively~\cite{newsedits,re3,versionrag,ledger,nepatec}. To guide fact comparison and document introduction, we record direct predecessor links between versions, reports, or assessment stages. For subsequent fact grounding, we retain each document's text and order in identifiable passages.

\subsection{Fact grounding}
\label{sec:document_graph}
To provide traceable document premises for later tasks, we express each fact as an independently checkable statement grounded in original document passages. Statements retain the original attribution, modality, and applicable conditions. When a fact recurs, we retain independent support in each relevant document so evidence can be drawn from documents already introduced at the request.

Applying these grounding principles, we first compare the same entity's property or relation across documents linked by predecessor relations, recording substantive changes as document evolution relations. We then ground rules, conditions, and other facts that later requests may use independently or alongside changed information. After review against the original documents, the facts, support links, and evolution relations form the Document Graph $G_D$ for workflow planning and Support Graph construction.

\subsection{Overall workflow planning}
\label{sec:workflow_planning}
To organize continuing work, we use $D$ and $G_D$ to develop an overall plan $P$ with objectives and topics for successive sessions. Documents are scheduled for introduction when their facts or changes advance the session objective, following predecessor order. The plan also allows sessions to progress through user updates or reuse of earlier results without introducing new documents.

Within this workflow, we allocate target task counts by category and response type according to the support and conditions each requires, preserving the same overall task composition across scenarios. Before detailed session design, we review the plan for cross-session continuity and task feasibility.

\subsection{Dialogue and task construction}
\label{sec:dialogue_task_construction}
Following $P$, we design each session using earlier dialogue, the accumulated Support Graph, and document facts from $G_D$. The design places document introductions, user updates, and reuse of earlier results in rounds that advance the session objective. For each task $\tau$, it identifies the necessary, applicable premises $A_\tau$ available at request $q_t$. Response content and type depend on the request, what these premises establish, and any unresolved user choices, conflicting applicable claims, or factual gaps.

After reviewing the design, we generate dialogue and check it against the design and original sources. For dialogue-grounded and jointly grounded tasks, this review verifies that prior user information remains necessary after accounting for request restatements and earlier replies.

Using the reviewed design and dialogue, we extend the Support Graph $G$ to preserve original support through updates and reuse. The graph inherits introduced document facts, their support links, and evolution relations from $G_D$. It records user conditions, choices, and observations as dialogue facts grounded in original user turns, linking prior and revised facts when updates occur. Support links record each intermediate conclusion's derivation from at least two jointly necessary premises, and task links connect each task $\tau$ to its premises $A_\tau$. Together, these connections preserve paths to original turns and passages when document facts are restated in dialogue or conclusions reused. The reviewed dialogue $H$ and updated graph then guide the next session.

\subsection{Task annotation and validation}
\label{sec:validation_assembly}
Using the premises $A_\tau$ and their dependencies in $G$, we check each task's categories against the definitions in Section~\ref{sec:task_categories}. For the required response, we trace the necessary support through $G$ to original turns and document passages available at $q_t$, including updates needed to establish applicability. The annotation records the final reviewed assistant reply as reference content, the required response type, and the traced sources as the default complete evidence set $C_t^0$. We verify that $C_t^0$ and each accepted alternative independently justify the required response under the same request conditions, with every entry necessary. All tasks underwent the construction checks detailed in Appendix~\ref{sec:construction_details}, with revision or reconstruction where needed. A final human review led to manual revisions of requests or reference responses for 262 tasks.

Through this process, we construct tasks that assess how assistants use evolving dialogue and documents in continuing work to fulfill requests, choose appropriate responses, and cite original supporting evidence. The final benchmark contains original documents, multi-session dialogue, annotated tasks, and the Support Graph. Its 1,000 tasks span five scenarios and include 700 jointly grounded tasks. Table~\ref{tab:dataset} reports corpus statistics, with detailed task counts in Appendix~\ref{sec:task_composition}.

\section{Experiments}
\label{sec:experiments}

\begin{table*}[t]
\centering
\small
\caption{Baseline performance on DuSE-Bench (\%). Grouped columns report content accuracy by grounding source and state dependence (700 jointly grounded tasks). Overall columns report content accuracy (Acc.), response-type Macro-F1 (M-F1), and source attribution F1 across all 1,000 tasks. Dlg./Doc. denote dialogue/document. \textbf{Bold} and \underline{underlining} indicate the best and second-best scores per column.}
\label{tab:main}
\begin{tabular}{ll|*{3}{w{c}{20pt}}|*{4}{w{c}{20pt}}|*{3}{w{c}{20pt}}}
\toprule
\multirow{2}{*}[-2pt]{\textbf{ID}} & \multirow{2}{*}[-2pt]{\textbf{Setting}} & \multicolumn{3}{c|}{\textbf{Grounding source}} & \multicolumn{4}{c|}{\textbf{State dependence (Joint)}} & \multicolumn{3}{c}{\textbf{Overall}} \\
\cmidrule(lr){3-5}\cmidrule(lr){6-9}\cmidrule(lr){10-12}
& & \textbf{Dlg.} & \textbf{Doc.} & \textbf{Joint} & \textbf{Static} & \textbf{Dlg.} & \textbf{Doc.} & \textbf{Dual} & \textbf{Acc.} & \textbf{M-F1} & \textbf{F1} \\
\midrule
\rowcolor[HTML]{F0F4F8}
\multicolumn{12}{c}{\textbf{Long context}} \\
A1 & LC-4K & 59.3 & 47.3 & 54.6 & 52.7 & \underline{64.0} & 50.0 & 55.0 & 54.20 & 39.16 & 26.04 \\
A2 & LC-32K & 56.0 & 56.7 & 55.7 & \underline{55.0} & 56.0 & 50.0 & \underline{59.5} & 55.90 & \underline{39.50} & 23.61 \\
\midrule
\rowcolor[HTML]{F0F4F8}
\multicolumn{12}{c}{\textbf{Retrieval}} \\
A3 & Dlg-RAG & 66.7 & 42.7 & 50.3 & 50.3 & 50.0 & 48.0 & 51.5 & 51.60 & 36.45 & 31.69 \\
A4 & Doc-RAG & 26.7 & 60.7 & 32.3 & 32.7 & 40.0 & 27.0 & 30.5 & 35.70 & 19.55 & 8.35 \\
A5 & Joint-RAG & 64.0 & 58.7 & 52.0 & 53.7 & 56.0 & 43.0 & 52.0 & 54.80 & 36.37 & \textbf{33.69} \\
\midrule
\rowcolor[HTML]{F0F4F8}
\multicolumn{12}{c}{\textbf{Raw dialogue history}} \\
A6 & RH + Doc-RAG & 66.0 & 62.0 & 54.4 & 51.3 & 61.0 & 49.0 & 58.5 & 57.30 & 39.39 & 26.94 \\
A7 & FH + Doc-RAG & 60.7 & \textbf{63.3} & 54.9 & 54.3 & 59.0 & \textbf{61.0} & 50.5 & 57.00 & \textbf{41.85} & 21.04 \\
A8 & RH + Joint-RAG & \underline{68.0} & 58.0 & \textbf{56.7} & 53.7 & 59.0 & \underline{56.0} & \textbf{60.5} & \underline{58.60} & 33.01 & \underline{31.93} \\
\midrule
\rowcolor[HTML]{F0F4F8}
\multicolumn{12}{c}{\textbf{Dialogue memory}} \\
A9 & MemGPT + Doc-RAG & \textbf{70.7} & \underline{62.7} & \underline{56.4} & \textbf{56.3} & \textbf{67.0} & 49.0 & 55.0 & \textbf{59.50} & 33.30 & 29.71 \\
A10 & A-Mem + Doc-RAG & 46.0 & 53.3 & 46.3 & 46.7 & 53.0 & 38.0 & 46.5 & 47.30 & 32.40 & 31.90 \\
A11 & Mem0 + Doc-RAG & 32.7 & 55.3 & 34.3 & 39.0 & 38.0 & 27.0 & 29.0 & 37.20 & 23.53 & 14.94 \\
\bottomrule
\end{tabular}
\end{table*}

Our experiments address four research questions about using evolving dialogue and documents:
\begin{itemize}[leftmargin=1.5em]
  \item \textbf{RQ1:} How does task completion vary by grounding source and state dependence?
  \item \textbf{RQ2:} How well do methods choose appropriate response types and cite original supporting evidence?
  \item \textbf{RQ3:} Why does joint retrieval miss document evidence, and how do retrieval strategies affect coverage and response performance?
  \item \textbf{RQ4:} How does the benefit of replacing one source's evidence depend on support from the other?
\end{itemize}

\subsection{Experimental setup}
\label{sec:experimental_setup}
We compare four approaches to supplying dialogue and document information. All configurations in the main experiments use GPT-4.1-mini with the same response prompt, following common evaluation practice in retrieval and memory benchmarks~\cite{rhelm,mtrag,membench}. Appendix~\ref{sec:implementation} details input construction and implementation. Additional LLMs are compared in Appendix~\ref{sec:model_performance}.

\textbf{Long context.}
Long context (LC) retains the latest dialogue and documents in chronological order within 4,000-token (LC-4K, A1) or 32,000-token (LC-32K, A2) windows.

\textbf{Retrieval.}
Retrieval-augmented generation (RAG)~\cite{rag} uses dense ranking to retrieve up to ten items. Dialogue-only RAG (Dlg-RAG, A3) retrieves complete dialogue rounds, while document-only RAG (Doc-RAG, A4) retrieves chunks from original passages. Joint RAG (Joint-RAG, A5) ranks both item types together.

\textbf{Raw dialogue history.}
Recent history (RH) retains the latest complete rounds within 4,000 dialogue tokens, while full history (FH) retains all prior dialogue. RH + Doc-RAG (A6) and FH + Doc-RAG (A7) add these histories to A4's document inputs, preserving their text and order. RH + Joint-RAG (A8) adds RH while preserving all A5 materials.

\textbf{Dialogue memory.}
MemGPT + Doc-RAG (A9)~\cite{memgpt}, A-Mem + Doc-RAG (A10)~\cite{amem}, and Mem0 + Doc-RAG (A11)~\cite{mem0} supply dialogue memories alongside A4's unchanged document inputs.

\subsection{RQ1: Task completion by grounding source and state~dependence}
\label{sec:experiment_a}
\textbf{Joint task completion.}
Across the evaluated long-context, retrieval, and memory configurations, joint-task content accuracy reaches at most 56.7\pct{} (Table~\ref{tab:main}). For each configuration, joint-task accuracy is also below accuracy on its better-performing single-source category, although the gap varies. For example, Joint-RAG reaches 64.0\pct{} on dialogue-grounded tasks and 58.7\pct{} on document-grounded tasks, compared with 52.0\pct{} on jointly grounded tasks. MemGPT + Doc-RAG likewise reaches 70.7\pct{} on dialogue-grounded tasks but 56.4\pct{} on jointly grounded tasks. Jointly grounded requests may require document facts to be interpreted under earlier user conditions that are not fully restated in the current request. Consistent with this requirement, adding RH to Doc-RAG improves joint-task accuracy by 22.1 pp with document inputs fixed. The gain may reflect recent dialogue supplying conditions or facts missing from the document inputs.

\textbf{Uneven handling of source changes.}
Within jointly grounded tasks, document-state accuracy is lower than dialogue-state accuracy in ten of the eleven configurations. This pattern appears in both long-context settings and Joint-RAG, as well as the three memory methods, whose within-method gaps range from 11 to 18 pp. Although the relevant changes occur in documents, retained user conditions still govern how the changed facts should be used. The required response can therefore change even when those user conditions remain unchanged. FH + Doc-RAG reverses this ordering, reaching 61.0\pct{} on document-state tasks and 59.0\pct{} on dialogue-state tasks. Its retrieved document inputs are identical to those supplied to RH + Doc-RAG and the memory configurations. Differences in document-state accuracy suggest that dialogue input also matters when handling document changes.

\textbf{Dual-state task completion.}
Relatively strong single-state performance does not consistently extend to dual-state tasks. For example, FH + Doc-RAG reaches 50.5\pct{} on dual-state tasks, below its accuracy on both single-state categories. MemGPT + Doc-RAG reaches 55.0\pct{}, 12.0 pp below its dialogue-state accuracy, although still above its document-state accuracy. In dual-state tasks, the assistant must identify the applicable user conditions and document facts and use them together to determine the required response. In particular, an earlier condition may remain topically relevant even after it has been superseded. RH + Joint-RAG shows a different profile, reaching 60.5\pct{} on dual-state tasks, above both of its single-state accuracies. Even so, it fails to apply the explicit phase update in the software-release task illustrated in Figure~\ref{fig:dialogue-example}. Although its input moves warning work from the pilot phase to release handoff, it identifies the phase as pilot and recommends sandbox notes.

In summary, joint-task accuracy is below each configuration's best single-source accuracy. Within joint tasks, document-state accuracy is generally lower than dialogue-state accuracy, and single-state strengths do not consistently extend to dual-state tasks. These patterns indicate persistent difficulty in combining applicable dialogue conditions and document facts.

\begin{figure*}[t]
\centering
\includegraphics[width=\textwidth]{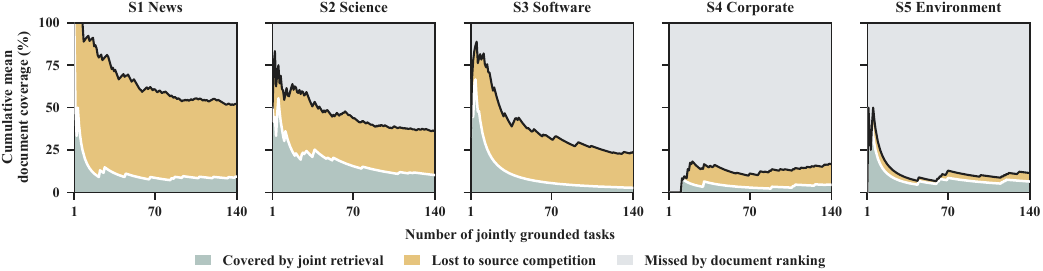}
\Description{Five panels show cumulative mean document coverage on 700 jointly grounded tasks, with 140 per scenario. The horizontal axis counts jointly grounded tasks in request order. Coverage is the fraction of default-annotated document passages represented by retrieved chunks. The white boundary shows joint Top-10 coverage; the black boundary shows document-only Top-10 coverage. Teal is the covered fraction under joint retrieval, gold is the additional coverage available under document-only retrieval, and light gray is the fraction still uncovered by document-only retrieval. All values are averaged over tasks encountered so far, and the three areas sum to 100 percent.}
\caption{Cumulative mean document coverage and retrieval gaps on 700 jointly grounded tasks (140 per scenario). White and black curves show Top-10 coverage under Joint-RAG (A5) and document-only retrieval, respectively. Teal, gold, and gray denote joint coverage, loss to source competition, and the document-ranking gap. Coverage is the fraction of default-annotated document passages represented by retrieved chunks and is averaged over tasks encountered so far in request order.}
\label{fig:a5_joint_tasks}
\vspace{-5pt}
\end{figure*}

\subsection{RQ2: Response choice and source attribution}
\label{sec:response_choice_attribution}
\textbf{Answering despite unresolved issues.}
Beyond content completion, all eleven baselines achieve higher F1 for answers than for clarification requests, conflict reports, or abstention. Across these categories, Joint-RAG chooses to answer instead on 64--88\pct{} of tasks. Recording that a user choice remains unresolved does not necessarily lead to a clarification request. In a science task, A-Mem retrieves a note requiring it to ask whether the recruitment worksheet is for test development or intervention. Although its response acknowledges that the study part remains unselected, it gives worksheet instructions rather than asking the user to choose.

\textbf{Source attribution for content-correct responses.}
Even responses judged content-correct frequently lack complete source attribution. Across baselines, 79--92\pct{} of these responses leave some annotated evidence uncited. Among Joint-RAG's 364 content-correct responses to jointly grounded tasks, source attribution F1 averages 46.3\pct{} for dialogue but only 4.3\pct{} for documents. In a news task, Joint-RAG correctly concludes that the reported total of 365 exceeds the user's threshold of 350. Yet it cites only the user's rule and an earlier assistant reply repeating the total, leaving the original document passage uncited.

\textbf{Different effects of additional dialogue.}
Additional dialogue can improve one metric while lowering another. For example, adding RH to Joint-RAG raises overall content accuracy by 3.8 pp but lowers response-type Macro-F1 by 3.4 pp. Clarification F1 falls from 33.9\pct{} to 19.3\pct{}. Extending RH to FH alongside Doc-RAG instead raises Macro-F1 from 39.4\pct{} to 41.8\pct{}, while content accuracy changes little. Improved conflict recognition accounts for much of this gain. Yet RH already contains every default-annotated dialogue turn for each of the seven conflict tasks newly classified correctly under FH. Broader history may therefore help interpret already-retained claims in context. However, the same history extension lowers overall source attribution F1 from 26.9\pct{} to 21.0\pct{}.

Overall, methods often choose to answer unresolved requests, while content-correct responses frequently lack complete attribution to original evidence. Additional dialogue can improve one metric while lowering another, supporting separate evaluation of content completion, response choice, and source attribution.

\subsection{RQ3: Document evidence recovery in joint retrieval}
\label{sec:experiment_b}
\textbf{Document retrieval bottlenecks.}
To examine the document support available to the model, we analyze Joint-RAG (A5) on the 700 jointly grounded tasks. Document coverage is the fraction of default-annotated passages represented by retrieved chunks, averaged over tasks. A5 achieves only 6.6\pct{} document coverage, compared with 28.0\pct{} under document-only Top-10 retrieval. The difference measures coverage lost to dialogue competition, while the fraction still uncovered by document-only retrieval is the ranking gap. Most of the uncovered fraction persists under document-only retrieval. Figure~\ref{fig:a5_joint_tasks} shows cumulative mean coverage generally declining in News, Science, and Software, but remaining low with fluctuations in Corporate and Environment.

\textbf{Retrieval strategies and response performance.}
Guided by these bottlenecks, we compare three interventions targeting capacity, source allocation, and ranking. B1 expands the joint Top-10 to 20 while retaining A5's original items; B2 allocates five slots to each source; B3 combines dense and sparse rankings within the original ten-slot budget. Table~\ref{tab:b_interventions} reports changes in source-specific coverage and content accuracy on the 700 jointly grounded tasks, with dialogue coverage computed as the fraction of default-annotated turns represented in the input. The diagnostic counts record zero-to-positive document-coverage recoveries and positive-to-zero losses; Appendix~\ref{sec:retrieval_diagnostics} defines their bottleneck grouping. Overall response metrics use all 1,000 tasks and are reported in Appendix Table~\ref{tab:intervention_results}.
\begin{table}[t]
\centering
\caption{Retrieval interventions (B1--B3) relative to Joint-RAG (A5) on 700 jointly grounded tasks. $\Delta$ columns report changes in evidence coverage and content accuracy (Acc.) in percentage points (pp). Task transitions list zero-to-positive document-coverage recoveries for source-competition/document-ranking cases followed by positive-to-zero losses shown as negative counts. Coverage and case definitions are provided in Appendix~\ref{sec:input_coverage}.}
\label{tab:b_interventions}
\begingroup
\small
\setlength{\tabcolsep}{3pt}
\renewcommand{\arraystretch}{1.05}
\setlength{\aboverulesep}{0.3ex}
\setlength{\belowrulesep}{0.45ex}
\begin{tabular}{lrrrr}
\toprule
\multirow{2}{*}[-2pt]{\textbf{Setting}} & \multirow{2}{*}[-2pt]{\textbf{\shortstack{Task\\transitions}}} & \multicolumn{2}{c}{\textbf{$\Delta$ Coverage (pp)}} & \multirow{2}{*}[-2pt]{\textbf{$\Delta$ Acc. (pp)}} \\
\cmidrule(lr){3-4}
& & \textbf{Document} & \textbf{Dialogue} & \\
\midrule
B1 & $29/2/0$ & $+4.23$ & $+7.73$ & $+2.29$ \\
B2 & $111/0/-3$ & $+13.07$ & $-8.79$ & $-0.57$ \\
B3 & $22/20/-5$ & $+4.74$ & $+3.98$ & $+1.71$ \\
\bottomrule
\end{tabular}
\endgroup
\end{table}

Among tasks with zero A5 document coverage, B1 mainly restores positive coverage in source-competition cases, with few document-ranking recoveries. Most document-ranking cases have no reference chunk in the document-only Top-20, putting them beyond B1's expanded joint budget. Both sources' coverage rises and overall content accuracy improves by 2.3 pp, but response-type Macro-F1 falls. B2 achieves the largest document-coverage gain of 13.1 pp, but dialogue coverage falls by 8.8 pp. All four overall response point estimates decline. In a software task, B2 retrieves the relevant WebAssembly passage but omits the update selecting a streaming fixture with an invalid WebAssembly response. Its reply uses the earlier buffer fixture, whereas A5 retains the update and correctly describes the selected streaming fixture. B3 instead revises ranking at the original capacity, restoring positive document coverage on 20 document-ranking cases, compared with two for B1. Relative to A5, both sources' coverage and all four overall response point estimates improve.

In summary, document ranking dominates A5's coverage gap. Larger budgets can leave ranking gaps unresolved, while source quotas can displace necessary dialogue support. Retrieval gains should therefore be assessed using both sources' coverage and overall response performance.

\subsection{RQ4: Interaction between dialogue and document evidence}
\label{sec:experiment_c}
\begin{figure}[!t]
\centering
\includegraphics[width=0.92\columnwidth]{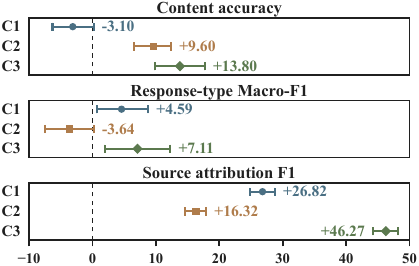}
\Description{Three panels show changes in content accuracy, response-type Macro-F1, and source attribution F1 relative to Joint-RAG (A5). In each panel, C1 substitutes annotated original dialogue turns, C2 original document passages, and C3 both. Points show paired changes in percentage points, with pointwise 95 percent bootstrap intervals obtained by resampling triggering sessions within each scenario.}
\caption{Paired performance changes from Joint-RAG (A5) after gold-evidence substitution on all 1,000 tasks, in percentage points (pp). C1 replaces dialogue, C2 documents, and C3 both with annotated original evidence. Bars show pointwise 95\% bootstrap intervals from session resampling within scenarios.}
\label{fig:oracle_sources}
\vspace{-5pt}
\end{figure}

The retrieval results raise a further question: does the benefit of supplying reference evidence from one source depend on support from the other? We examine this using matched evidence substitutions in A5. C1 replaces retrieved dialogue with annotated original turns, C2 replaces document chunks with annotated original passages, and C3 replaces both. For each task, the annotated evidence from a source is identical across configurations, while single-source substitutions preserve A5's materials and order for the other source. Each substitution also removes the original retrieved materials from that source, potentially excluding useful restatements or surrounding context. Figure~\ref{fig:oracle_sources} reports paired changes on all 1,000 tasks.

\textbf{Single-source substitutions.}
Accuracy and response choice shift in opposite directions. C1 raises response-type Macro-F1 but lowers the content-accuracy point estimate, whereas C2 raises content accuracy by 9.6 pp but lowers the Macro-F1 point estimate. C2 also increases answer predictions, suggesting that stronger factual support does not reliably resolve whether a direct answer is appropriate. Both substitutions improve source attribution F1, consistent with making original supporting evidence explicit.

\textbf{Conditional gains from joint support.}
C3 achieves the highest point estimates on all four aggregate metrics among A5 and C1--C3, reaching 68.6\pct{} content accuracy (Appendix Table~\ref{tab:intervention_results}). The effect of dialogue substitution changes with the document input: it lowers content accuracy by 3.1 pp with A5's documents but raises it by 4.2 pp with annotated documents. To quantify how much the joint gain exceeds the sum of the single-source gains, we define the interaction for score $S$ as
\begin{equation}
\label{eq:source_interaction}
 I_S=S(\mathrm{C3})-S(\mathrm{C2})-S(\mathrm{C1})+S(\mathrm{A5}).
\end{equation}
The content-accuracy interaction is $+7.3$ pp (95\% bootstrap interval: 3.4--11.2). Its point estimate is also positive for response-type Macro-F1 and source attribution F1.

A news task illustrates how the two sources constrain the required response. The updated brief calls for concerns about future grievance reporting and aims for greater transparency. Because A5 retrieves no document passages, C1 receives the current requirements without the supporting facts and merely restates the brief. C2 receives the relevant passages but retains A5's dialogue without the update; its response mixes the earlier retrospective scope with the newer requirements and still omits Aluko's specific concerns and aims. With both the updated brief and the original passages, C3 reports her concern that her experience may deter other players from speaking out and her call for greater transparency.

In summary, the benefit of replacing one source's evidence depends on support from the other. The remaining content and attribution errors under C3 show that supplying annotated evidence does not ensure correct responses or complete citation. These findings support coordinating dialogue memory and document retrieval to apply current user conditions and document facts together.

\section{Conclusion}
We present \bench{} to evaluate joint reasoning over evolving dialogue and documents in persistent Web information assistants. Its 1,000 tasks span five multi-session scenarios where user requirements govern document use and earlier conclusions support subsequent work. An evolving Support Graph guides construction and traces updated facts and reused conclusions to original turns and passages. Annotations distinguish grounding source from state dependence and enable separate evaluation of response content, response type, and source attribution.

Experiments across 11 baseline configurations show persistent difficulty in combining applicable user requirements and document facts. Gains in content accuracy can coincide with poorer response choice, while content-correct responses frequently lack complete attribution to original evidence. Matched evidence substitutions further show that the effect of replacing one source's evidence depends on support from the other. Together, these findings motivate coordinating dialogue memory and document retrieval to apply evidence under the conditions governing each response. \bench{} provides a testbed for studying this coordination while separately assessing whether responses fulfill requests, choose appropriate response types, and cite original supporting evidence.

\bibliographystyle{template/ACM-Reference-Format}
\begingroup
\def\showURL#1{\unskip}
\def\href#1#2{}
\renewcommand{\showeprint}[2][arxiv]{arXiv:#2}
\bibliography{sections/references}
\endgroup
\appendix
\section{Benchmark Details}
\label{sec:coverage}
\subsection{Task composition}
\label{sec:task_composition}
The five scenarios share the same task allocation by grounding source and state dependence, with 200 tasks in each (Table~\ref{tab:profiles}). Each contains 140 \policy{answer} tasks and 20 each requiring \policy{clarification}, \policy{conflict reporting}, or \policy{abstention}. Across the benchmark, answer tasks account for 61\pct{} of jointly grounded tasks and 91\pct{} of single-source tasks. The 140 jointly grounded tasks per scenario are equally divided between direct and chained integration.
\begin{table}[ht]
\centering
\begingroup
\captionsetup{font=normalsize,width=\linewidth}
\caption{DuSE-Bench task counts by grounding source and state dependence, pooled across five scenarios.}
\label{tab:profiles}
\begin{threeparttable}
\small
\setlength{\tabcolsep}{5pt}
\renewcommand{\arraystretch}{1.05}
\setlength{\aboverulesep}{0.3ex}
\setlength{\belowrulesep}{0.45ex}
\begin{tabular}{l*{5}{c}}
\toprule
\multirow{2}{*}[-2pt]{\textbf{\shortstack{Grounding\\source}}} & \multicolumn{4}{c}{\textbf{State dependence}} & \multirow{2}{*}[-2pt]{\textbf{Total}} \\
\cmidrule(lr){2-5}
& \textbf{Static} & \textbf{Dialogue} & \textbf{Document} & \textbf{Dual} & \\
\midrule
Dialogue & 50 & 100 & -- & -- & \textbf{150} \\
Document & 50 & -- & 100 & -- & \textbf{150} \\
Joint & 300 & 100 & 100 & 200 & \textbf{700} \\
\midrule
\rowcolor[HTML]{F0F4F8}
\textbf{Total} & \textbf{400} & \textbf{200} & \textbf{200} & \textbf{200} & \textbf{1,000} \\
\bottomrule
\end{tabular}
\begin{tablenotes}[flushleft]
\footnotesize
\item Dashes indicate inapplicable combinations.
\end{tablenotes}
\end{threeparttable}
\endgroup
\end{table}

\textbf{Chained-support example.}
\label{sec:worked_evidence_paths}
An earlier software task combines the rule that all selected criteria must pass with an updated callback criterion, establishing that approval requires exactly two invocations in the selected callback test. A later task uses a document passage describing count tracking and explicit verification to specify how to check this requirement. Its complete evidence set contains the two original user turns and the document passage.

\subsection{Validation criteria}
\label{sec:construction_details}
\textbf{Document facts and changes.}
Document review verifies predecessor links and checks that original passages support each fact and its qualifications. Changes are verified by comparing the same entity's property or relation across linked documents, with supporting evidence from the respective documents. First mention or omission alone does not establish an addition or withdrawal.

\textbf{Task dependencies.}
Dialogue-grounded and jointly grounded tasks require prior user information after accounting for the current request and restated document facts. State dependence is tested by restoring one source's prior state while holding the other fixed and checking whether the required response changes. Chained-task review checks the earlier conclusion's derivation from jointly necessary premises and its contribution to the later response.

\textbf{Reference responses and evidence.}
Reference content and response type are checked against the core request under the applicable version, time, scope, and priority rules. Evidence is restricted to original turns and passages available at the request, including necessary updates. Each complete set must jointly justify the required response, with every entry necessary. Alternatives must independently meet these criteria for the same response under the same request conditions and preserve the task categories.

\subsection{Benchmark comparison criteria}
\label{sec:comparison_definitions}
Capability labels in Table~\ref{tab:benchmark_comparison} follow the task definitions and evaluation protocols documented in the original reports and releases~\cite{locomo,longmemeval,versionrag,tide,mtrag,memorydocdataset,rhelm}. Document lengths follow released source units, including paragraph records in mtRAG and whole files in RHELM. \textbf{Cross-session dialogue} requires tasks that depend on information from earlier original sessions. \textbf{Document evolution} requires task-relevant factual changes across versions, successive reports, or recorded document modifications. \textbf{Joint reasoning} requires both dialogue and document information to determine a response, including document identification or application of user conditions. \textbf{Response-type evaluation} requires an explicit metric or judging rule for response choice, such as whether to answer or abstain. \textbf{Source attribution} requires separate evaluation of model-supplied source identifiers or citations; reference annotations, retrieval scores, and answer faithfulness alone do not qualify.

\section{Experimental Protocol}
\label{sec:implementation}
\subsection{Input configurations}
\label{sec:input_assembly}
\textbf{Shared settings.}
The main response model and content judge use GPT-4.1-mini at temperature zero. Each configuration follows the same recorded dialogue, with history and memory carried across sessions within each scenario. At each request, available documents comprise those introduced up to that point. Materials retain their original source identifiers, and gold-evidence inputs contain complete original turns or passages from $C_t^0$.

\begin{table}[ht]
\centering
\begingroup
\captionsetup{font=normalsize,width=\linewidth}
\caption{Scoring rubric for response content quality (0--5).}
\label{tab:content_rubric}
\begin{threeparttable}
\small
\setlength{\tabcolsep}{4pt}
\renewcommand{\arraystretch}{1.05}
\setlength{\aboverulesep}{0.3ex}
\setlength{\belowrulesep}{0.45ex}
\begin{tabular}{>{\centering\arraybackslash}m{26pt}>{\raggedright\arraybackslash}m{\dimexpr0.95\columnwidth-26pt-4\tabcolsep\relax}}
\toprule
\rowcolor[HTML]{F0F4F8}
\textbf{Score} & \multicolumn{1}{c}{\textbf{Quality criterion}} \\
\midrule
5 & Complete, accurate fulfillment. \\
4 & Correct core results and essential qualifications, with minor omissions or peripheral inaccuracies. \\
3 & Substantial correct content with an essential component missing or wrong. \\
2 & Some concrete correct content, with most essential content missing or wrong, or an incorrect central conclusion undermining the result. \\
1 & Marginally useful correct information or recognition of the topic. \\
0 & No relevant correct content. \\
\bottomrule
\end{tabular}
\end{threeparttable}
\endgroup
\end{table}

\begin{table*}[!t]
\centering
\small
\captionsetup{width=\linewidth}
\caption{LLM performance on DuSE-Bench (\%). Grouped columns report content accuracy by grounding source and state dependence (700 jointly grounded tasks). Overall columns report content accuracy (Acc.), response-type Macro-F1 (M-F1), and source attribution F1 across all 1,000 tasks. Dlg./Doc. denote dialogue/document. \textbf{Bold} and \underline{underlining} indicate the best and second-best scores per column.}
\label{tab:model_performance}
\begin{tabular}{ll|*{3}{w{c}{20pt}}|*{4}{w{c}{20pt}}|*{3}{w{c}{20pt}}}
\toprule
\multirow{2}{*}[-2pt]{\textbf{Setting}} & \multirow{2}{*}[-2pt]{\textbf{Model}} & \multicolumn{3}{c|}{\textbf{Grounding source}} & \multicolumn{4}{c|}{\textbf{State dependence (Joint)}} & \multicolumn{3}{c}{\textbf{Overall}} \\
\cmidrule(lr){3-5}\cmidrule(lr){6-9}\cmidrule(lr){10-12}
& & \textbf{Dlg.} & \textbf{Doc.} & \textbf{Joint} & \textbf{Static} & \textbf{Dlg.} & \textbf{Doc.} & \textbf{Dual} & \textbf{Acc.} & \textbf{M-F1} & \textbf{F1} \\
\midrule
\multirow{4}{*}{Joint-RAG} & GPT-4.1-mini & 64.0 & \underline{58.7} & 52.0 & 53.7 & 56.0 & 43.0 & 52.0 & 54.80 & 36.37 & 33.69 \\
 & GPT-5.1 & \textbf{73.3} & \textbf{69.3} & \textbf{65.9} & \textbf{69.0} & \textbf{63.0} & \textbf{61.0} & \textbf{65.0} & \textbf{67.50} & \underline{45.54} & \textbf{34.39} \\
 & Claude Sonnet 4.5 & \underline{68.7} & 58.0 & 55.1 & \underline{57.7} & 55.0 & 52.0 & 53.0 & 57.60 & \textbf{53.09} & \underline{33.81} \\
 & Qwen2.5-7B-Instruct & 45.3 & 42.7 & 33.9 & 36.3 & 32.0 & 26.0 & 35.0 & 36.90 & 38.00 & 19.10 \\
\midrule
\multirow{2}{*}{RH + Joint-RAG} & GPT-4.1-mini & 68.0 & 58.0 & \underline{56.7} & 53.7 & \underline{59.0} & \underline{56.0} & \underline{60.5} & \underline{58.60} & 33.01 & 31.93 \\
 & Qwen2.5-7B-Instruct & 46.0 & 35.3 & 31.7 & 35.3 & 23.0 & 30.0 & 31.5 & 34.40 & 30.38 & 13.90 \\
\bottomrule
\end{tabular}
\end{table*}

\begin{table}[!t]
\centering
\captionsetup{font=normalsize,width=\linewidth}
\caption{Response performance under retrieval interventions (\mbox{B1--B3}) and gold-evidence substitutions (\mbox{C1--C3}) (\%). A5 is the Joint-RAG baseline. Columns report content accuracy (Acc.), content quality (Qual.), response-type Macro-F1 (M-F1), and source attribution F1 across all 1,000 tasks. \textbf{Bold} and \underline{underlining} indicate the best and second-best scores per column within each group.}
\label{tab:intervention_results}
\begin{threeparttable}
\begingroup
\small
\setlength{\tabcolsep}{4pt}
\renewcommand{\arraystretch}{1.05}
\setlength{\aboverulesep}{0.3ex}
\setlength{\belowrulesep}{0.45ex}
\begin{tabular}{w{c}{24pt}*{4}{w{c}{\dimexpr(0.95\columnwidth-24pt-10\tabcolsep)/4\relax}}}
\toprule
\multirow{2}{*}[-2pt]{\textbf{ID}} & \multicolumn{2}{c}{\textbf{Content}} & \multirow{2}{*}[-2pt]{\textbf{M-F1}} & \multirow{2}{*}[-2pt]{\textbf{F1}} \\
\cmidrule(lr){2-3}
& \textbf{Acc.} & \textbf{Qual.} & & \\
\midrule
A5 & 54.80 & 70.70 & 36.37 & 33.69 \\
\midrule
\rowcolor[HTML]{F0F4F8}
\multicolumn{5}{c}{\textbf{Retrieval interventions}} \\
B1 & \textbf{57.10} & \textbf{72.72} & \underline{34.39} & \underline{35.06} \\
B2 & 53.40 & 69.70 & 33.12 & 33.62 \\
B3 & \underline{56.70} & \underline{72.12} & \textbf{38.29} & \textbf{36.83} \\
\midrule
\rowcolor[HTML]{F0F4F8}
\multicolumn{5}{c}{\textbf{Gold-evidence substitutions}} \\
C1 & 51.70 & 68.70 & \underline{40.96} & \underline{60.51} \\
C2 & \underline{64.40} & \underline{76.92} & 32.72 & 50.00 \\
C3 & \textbf{68.60} & \textbf{79.42} & \textbf{43.48} & \textbf{79.95} \\
\bottomrule
\end{tabular}
\endgroup
\end{threeparttable}
\end{table}

\textbf{Retrieval.}
RAG ranks complete dialogue rounds and document chunks against the current request. To form document chunks, each original passage is split into segments of up to 1,024 tokens with 128-token overlap. Dense retrieval uses cosine similarity with \texttt{text-\allowbreak{}embedding-\allowbreak{}3-small}, and BM25 provides sparse rankings~\cite{bm25}. B3 then fuses ten candidates from each ranking through reciprocal rank fusion~\cite{rrf} and retains ten items.

\textbf{Context and memory.}
LC retains the latest whole messages and passages within its full-input budget, including the prompt and request, whereas RH budgets dialogue alone. For memory configurations, MemGPT, A-Mem, and Mem0 process the recorded dialogue chronologically using GPT-4.1-mini for memory management.

\subsection{Scoring and statistical analysis}
\label{sec:metric_details}
\textbf{Content judging.}
The judge independently scores accuracy and quality using the request, reference content, and complete original turns and passages in $C_t^0$~\cite{mtrag,longmemeval}. Quality follows Table~\ref{tab:content_rubric} and is divided by five before averaging. Checks cover factual correctness and essential qualifications, including version, scope, and applicable conditions. For unresolved requests, checks cover the specific missing user condition, incompatible applicable claims, or factual gap together with any supported partial result.

\textbf{Statistical analysis.}
\label{sec:bootstrap_details}
We use 2,000 cluster bootstrap draws for paired comparisons. Within each scenario, each draw resamples the observed number of triggering sessions with replacement and includes all tasks from each sampled session. Shared draws preserve pairing across configurations. We then recompute metrics on pooled tasks and calculate differences and interactions, whose bootstrap distributions yield pointwise 95\pct{} percentile intervals.

\begin{samepage}
\subsection{Evidence coverage and retrieval diagnostics}
\label{sec:input_coverage}
\textbf{Evidence coverage.}
Coverage for source $k$ is $|C_t^{0,k}\cap R_t^k|/|C_t^{0,k}|$. Here, $C_t^{0,k}$ contains identifiers from the default complete evidence set and $R_t^k$ contains supplied original identifiers.
Scores are averaged over tasks whose default evidence set contains that source. Each original turn or passage counts once, with a document passage represented by any of its retrieved chunks.
\end{samepage}

\textbf{Document retrieval decomposition.}
\label{sec:retrieval_diagnostics}
For the same request and document pool, A5's joint and document-only Top-10 retrieval use identical dense scoring. Their document coverages therefore satisfy $0\leq J_t\leq D_t\leq1$. Figure~\ref{fig:a5_joint_tasks} shows $J_t$ covered jointly, $D_t-J_t$ lost to source competition, and $1-D_t$ missed by document ranking. Among tasks with $J_t=0$, $D_t>0$ identifies source-competition cases and $D_t=0$ document-ranking cases.

\begin{samepage}
\section{Performance across LLMs}
\label{sec:supplementary_experiments}
\label{sec:model_performance}
Table~\ref{tab:model_performance} compares LLMs with retrieved materials fixed within each configuration. Under Joint-RAG, all four models achieve lower accuracy on joint tasks than on either single-source category. GPT-5.1's joint-task accuracy is highest at 65.9\pct{}, but still below its dialogue-grounded accuracy of 73.3\pct{}. Within joint tasks, document-state accuracy is lower than dialogue-state accuracy for all four models. Joint reasoning thus remains challenging across these LLMs.

Across all tasks, GPT-5.1 exceeds GPT-4.1-mini in content accuracy by 12.7 pp, while source attribution F1 improves by only 0.7 pp. Claude Sonnet 4.5 achieves the highest response-type Macro-F1 (53.1\pct{}) despite lower content accuracy than GPT-5.1. Across the four Joint-RAG models, 82\pct{} to 89\pct{} of content-correct responses still leave part of the selected complete evidence set uncited. Adding RH raises content accuracy for GPT-4.1-mini but lowers it for Qwen2.5-7B-Instruct, while response-type Macro-F1 and source attribution F1 decline for both. Overall, these results reinforce separate evaluation of content completion, response choice, and source attribution because gains in one need not extend to the others.
\end{samepage}

\onecolumn
\clearpage
\section{Case Studies}
\label{sec:case_studies}
Both cases use GPT-4.1-mini. M and D denote dialogue and document evidence. Evidence and responses are excerpted.\par
\kern2pt\begingroup
\setlength{\fboxsep}{3pt}
\setlength{\fboxrule}{0.4pt}
\newcommand{\casemeta}[2]{\mbox{\fontsize{8.5}{10.5}\selectfont\textcolor{black!65}{Response type:} \textbf{#1.}\quad\textcolor{black!65}{Supporting evidence:} \textbf{#2.}}}
\newcommand{\caseprediction}[3]{\noindent\textbf{#1}\hfill\makebox[50pt][r]{\textcolor[HTML]{35485E}{\textbf{#2}}}\hspace{10pt}\makebox[145pt][l]{\textcolor[HTML]{35485E}{\textbf{#3}}}\par\nobreak\kern3pt}
\newcommand{\casetitle}[1]{\par\noindent{\setlength{\fboxsep}{2pt}\colorbox[HTML]{E4EAF0}{\parbox{\dimexpr\linewidth-2\fboxsep\relax}{\fontsize{10}{11}\selectfont\color[HTML]{35485E}\bfseries #1}}}\par\kern2pt}
\newcommand{\caseheading}[1]{\par\kern2pt\noindent{\setlength{\fboxsep}{1pt}\colorbox[HTML]{F0F4F8}{\makebox[\dimexpr\linewidth-2\fboxsep\relax][l]{\strut\textcolor[HTML]{35485E}{\textbf{#1}}}}}\par\nobreak\kern1pt}
\newcommand{\casefield}[1]{\par\kern2pt\noindent\textcolor[HTML]{35485E}{\textbf{#1.}}\enspace}
\newcommand{\caseevidence}[2]{\par\begingroup\leftskip=25pt\noindent\llap{\makebox[25pt][l]{\textcolor[HTML]{35485E}{\textbf{[#1]}}}}#2\par\endgroup\kern1pt}
\noindent\fcolorbox{black!20}{white}{%
\begin{minipage}{\dimexpr\textwidth-2\fboxsep-2\fboxrule\relax}
\fontsize{9}{10.2}\selectfont\setlength{\parindent}{0pt}
\casetitle{Case 1: Software task. Applying updated conditions to document facts}
\casefield{Question}
For the currently selected warning phase and active legacy status, what particular handoff action does our routing rule require? Read both current settings rather than hard-code the old pilot disposition.\par
\casefield{Gold Answer}
Create a tracked release retirement-review followup for the active Deprecated ERR\_REQUIRE\_ESM classification.\par
\noindent\casemeta{Answer}{[M1], [M2], [M4], [D1]}\par
\caseheading{Evidence}
\caseevidence{M1}{\textcolor{black!65}{User.} ``\ldots{} Separately, our legacy-catalog routing sends Experimental classifications to pilot-readiness review and Deprecated classifications to retirement review; pilot entries remain sandbox notes, while release-stage entries become tracked followups. \ldots{}''}
\caseevidence{M2}{\textcolor{black!65}{User.} ``\ldots{} T is deferred outside the current handoff and its older observations remain historical. For ERR\_REQUIRE\_ESM, replace the working reference with the reviewed boundary Node19.9.0 error catalogue to Node20.19.0 error catalogue. \ldots{}''}
\caseevidence{M3}{\textcolor{black!65}{Assistant.} ``\ldots{} Node 20.19 legacy reference to retirement review under its active Deprecated classification \ldots{}''}
\caseevidence{M4}{\textcolor{black!65}{User.} ``\ldots{} into the release handoff, replacing the pilot phase. Its classification actions now need tracked release followups under our retained routing rule, without claiming package success or permitting newly introduced deprecated helpers.''}
\caseevidence{M5}{\textcolor{black!65}{Assistant.} ``The warning work is now in the release phase, so classification actions require tracked followups. \ldots{}''}
\caseevidence{D1}{\textcolor[HTML]{35485E}{\textbf{Document (Node 20.19.0).}} ``\ldots{} Stability: 0 - Deprecated An attempt was made to \texttt{require()} an ES Module. \ldots{}''}
\caseheading{Method Predictions}
\caseprediction{A5: Joint-RAG.}{Answer}{[M1], [M4], [M5]}
``For the currently selected warning phase \textbf{(release)} and active legacy status \textbf{(Deprecated classification)}, the routing rule requires that the handoff action be a \textbf{tracked release followup} without claiming package success or permitting newly introduced deprecated helpers.''\par
\kern2pt
\caseprediction{A8: RH + Joint-RAG.}{Answer}{[M4], [M5]}
``For the currently selected warning phase \textbf{(pilot)} and active legacy status, the routing rule requires that ESM warning work entries remain as \textbf{sandbox notes} rather than tracked release followups, meaning no passing package result is claimed \ldots{}''\par
\kern2pt
\caseprediction{A9: MemGPT + Doc-RAG.}{Answer}{[M4], [M5]}
``\ldots{} the routing rule requires that classification actions be \textbf{tracked as release followups}, replacing the pilot phase \ldots{}''\par
\kern2pt
\casefield{Failure Analysis}
All methods retain [M3] and [M4] without retrieving [D1]. A8 still uses pilot and sandbox notes. A9 omits retirement review retained in [M3]. A5 identifies release and Deprecated but leaves [M2] and [D1] uncited.\par
\end{minipage}}\par
\kern6pt
\noindent\fcolorbox{black!20}{white}{%
\begin{minipage}{\dimexpr\textwidth-2\fboxsep-2\fboxrule\relax}
\fontsize{9}{10.2}\selectfont\setlength{\parindent}{0pt}
\casetitle{Case 2: News task. Response choice and original-source attribution}
\casefield{Question}
Does the corrected security total trigger the editor’s revised review warning?\par
\casefield{Gold Answer}
Yes; the exact reported total of 365 exceeds the revised threshold of 350.\par
\noindent\casemeta{Answer}{[M1], [M2], [D1]}\par
\caseheading{Evidence}
\caseevidence{M1}{\textcolor{black!65}{User.} ``Update the security-death and Sanjit age entries from the new Manipur snapshot. Keep the previous readings labelled by source version.''}
\caseevidence{M2}{\textcolor{black!65}{User.} ``\ldots{} Trigger the review warning when an exact reported security-death total is greater than 350.''}
\caseevidence{M3}{\textcolor{black!65}{Assistant.} ``Yes. The current exact 365-versus-365 pair meets the numeric criterion. Retain the earlier lower-bound extract \ldots{}''}
\kern1pt{\fontsize{9}{10.2}\selectfont\noindent\textit{Other cited dialogue:} [M4] numeric query; [M5] age question; [M6] age reply; [M7] current query.\par}\kern1pt
\caseevidence{D1}{\textcolor[HTML]{35485E}{\textbf{Document.}} ``\ldots{} It said 365 police and paramilitaries had been killed by the rebels between 2000 and October 2012. \ldots{}''}
\caseheading{Method Predictions}
\caseprediction{A5: Joint-RAG.}{Answer}{[M2], [M3]}
``Yes, the corrected security total of \textbf{365} triggers the editor’s revised review warning, as it exceeds the threshold of \textbf{350}.''\par
\kern2pt
\caseprediction{A8: RH + Joint-RAG.}{Answer}{[M2], [M3]}
``Yes; the corrected exact reported security-death total is \textbf{365}, which exceeds the editor's revised review warning threshold of \textbf{350}, so the warning is triggered.''\par
\kern2pt
\caseprediction{A9: MemGPT + Doc-RAG.}{Abstention}{[M3], [M4], [M5], [M6], [M7]}
``The provided dialogue and documents \textbf{do not contain information} about a corrected security total or whether it triggers the editor’s revised review warning.''\par
\kern2pt
\casefield{Failure Analysis}
A5 and A8 answer correctly from [M2] and [M3] without retrieving [D1]. Their citations omit [M1] and [D1]. A9 retrieves [D1] but abstains with threshold [M2] absent from memory. It also cites unrelated dialogue, leaving [D1] uncited.\par
\end{minipage}}\par
\endgroup

\end{document}